\documentclass[letterpaper]{article} 
\usepackage[draft]{aaai2026}  
\usepackage{times}  
\usepackage{helvet}  
\usepackage{courier}  
\usepackage[hyphens]{url}  
\usepackage{graphicx} 
\usepackage{natbib}  
\usepackage{caption} 
\usepackage{algorithm}
\usepackage{algorithmic}

\usepackage{amsmath}

\usepackage{cleveref}
\usepackage[utf8]{inputenc}   

\usepackage{newfloat}
\usepackage{listings}
\DeclareCaptionStyle{ruled}{labelfont=normalfont,labelsep=colon,strut=off} 
\lstdefinestyle{noLineNums}{
    basicstyle={\footnotesize\ttfamily},
    numbers=none,
    aboveskip=0pt,
    belowskip=0pt,
    showstringspaces=false,
    tabsize=2,
    breaklines=true,
}
\lstdefinestyle{noLineNumsUTF8}{
  style=noLineNums,
  inputencoding=utf8,
  extendedchars=true,
  literate=
    {^^e2^^80^^a2}{{\textbullet}}1   
    {-}{{--}}1            
    {—}{{---}}1           
    {'}{{'}}1             
    {…}{{\dots}}1         
}

\lstdefinestyle{lcgPrompt}{
  basicstyle      = \ttfamily,  
  columns         = fullflexible,
  breaklines      = true,             
  breakatwhitespace = false,           
  breakindent     = 0pt,              
  breakautoindent = false,            
  numbers         = none,
  frame           = single,
  framerule       = .4pt,
  framesep        = 6pt,
  literate=
    {^^e2^^80^^a2}{{\textbullet}}1   
    {-}{{--}}1            
    {—}{{---}}1           
    {'}{{'}}1             
    {…}{{\dots}}1         
}

\usepackage{fancyvrb}

\floatstyle{ruled}
\newfloat{listing}{tb}{lst}{}
\floatname{listing}{Listing}
\usepackage{booktabs} 
\usepackage{tabularx}
\newcommand{\model}[1]{\texttt{#1}}   

\usepackage{multirow}

\makeatletter

\newenvironment{promptbox}[1]{%
  \par\medskip
  \noindent\begin{quote}
    \setlength{\parindent}{0pt}%
    \setlength{\parskip}{4pt}%
    \textbf{#1}\par
    \hrule height 0.4pt\relax
    \vspace{4pt}%
    \begin{minipage}{0.97\linewidth}%
}{%
    \end{minipage}%
    \vspace{4pt}\hrule height 0.4pt\relax
  \end{quote}\medskip
}

\newenvironment{codebox}{%
  \par\medskip
  \noindent\begin{quote}
    \setlength{\parindent}{0pt}%
    \hrule height 0.4pt\relax
    \vspace{4pt}\small\ttfamily
}{%
    \vspace{4pt}\hrule height 0.4pt\relax
  \end{quote}\medskip
}

\usepackage{placeins}

\title{Prompt-Based One-Shot Exact Length-Controlled Generation with LLMs}

\author {
    Juncheng Xie, Hung-yi Lee
}
\affiliations{
    f07942150@ntu.edu.tw, hungyilee@ntu.edu.tw
}

\begin{document}

\maketitle


\begin{abstract}
Controlling the length of text produced by large language models (LLMs) remains challenging: models frequently overshoot or undershoot explicit length instructions because they cannot reliably keep an internal token count. We present a \emph{prompt‑based, one‑shot} strategy that compels an off‑the‑shelf LLM to generate \textbf{exactly} a desired number of tokens—words (English) or characters (Chinese)—without any fine‑tuning or iterative sampling.  
The prompt appends countdown markers and explicit counting rules so that the model "writes while counting." We evaluate on four settings: open‑ended generation (1–1000 tokens), XSUM summarization, \textbf{MT‑Bench‑LI instruction following}, and the LIFEBENCH equal‑length track.  
On MT‑Bench‑LI, strict length compliance with \texttt{gpt‑4.1} leaps from below 30 \% under naïve prompts to above 95 \% with our countdown prompt—surpassing the popular \emph{draft‑then‑revise} baseline—while judged answer quality is preserved. These results demonstrate that precise length control can be achieved through prompt engineering alone, offering a lightweight alternative to training‑ or decoding‑based methods.
\end{abstract}

\section{Introduction}
Length control is the ability of a text‑generation system to terminate at an \emph{exact} number of words, sentences, or tokens.  Precise control is critical for applications such as headline writing, one‑sentence summaries, constrained creative tasks in educational settings, or social‑media posts with hard character limits.  Output that exceeds or undershoots the limit is unusable.

Unfortunately, even today’s strongest large language models (LLMs) struggle with strict limits.  
Existing approaches tackle this gap by \emph{modifying the model}: retraining with extra length labels, adding length‑aware adapters such as countdown embeddings, or employing reinforcement learning to punish over‑ or under‑shooting~\cite{feng2023cotmystery}.  These methods can succeed but incur substantial compute, require access to model weights, and cannot be applied to proprietary black‑box models.  Heuristic decoding tricks (e.g., early stopping plus log‑probability rescaling) help but still do not guarantee perfect compliance, especially for longer passages.

We take a different view.  Counting is hard for transformers because self‑attention lacks an explicit accumulator.  Instead of pushing the model to learn an internal counter, we \emph{externalize} the counter through prompt engineering: we embed visible countdown markers and clear rules so the model’s job reduces to pattern completion.  Our approach needs no gradient updates, auxiliary reward models, or multiple inference passes, and it transfers unchanged to both proprietary APIs \emph{and} state‑of‑the‑art open‑source LLMs.

This paper therefore asks: \textbf{Can we realize \underline{exact} length control using only prompts?}  We introduce a simple strategy, show that it yields nearly perfect ( 99 \%) compliance across diverse target lengths, including passages of several hundred tokens, in both English and Chinese, demonstrate the same effect on leading open source models, and dissect why it works. Our contributions are summarized as follows:
\begin{itemize}
\item We propose a novel prompt engineering strategy to achieve exact-length generation with LLMs in a single pass, using countdown markers and explicit counting rules in the prompt.
\item We present extensive experiments in English (word-length control) and Chinese (character-length control) showing that our method vastly outperforms naive prompting in exact length compliance, across a range of model types and target lengths.
\item We analyze the limitations of current LLMs in counting and length awareness, and we provide a dedicated discussion on why LLMs struggle with these basic tasks and how our approach mitigates some of these issues.
\end{itemize}

\section{Related Work}

\subsection{Training-Based Approaches for Length Control}
Training‑based methods modify model parameters to hit target lengths. Hansel \cite{Song2025Hansel} adds countdown tokens so the model stops exactly on cue, while L1 \cite{Aggarwal2025L1} uses reinforcement learning to cap chain‑of‑thought steps. Further work trains length‑aware summarizers \cite{Miculicich2023Summarization,Yu2021LenAtten}, up‑weights the EOS loss \cite{Stergiadis2025EOS}, or applies length‑conditioned preference optimization \cite{Li2024LMPO}. Constrained RL decoding balances length and quality \cite{Chan2021CMDP}. These approaches achieve near‑exact control but require task‑specific fine‑tuning.

\subsection{Inference-Time Techniques}
Length can also be steered without retraining. Instruction‑tuned LLMs follow plain‑language length cues but remain approximate \cite{Zhou2023InstructCTG}. BB‑MH \cite{Gu2024BlackBox} enforces a hard upper bound by rejecting over‑long samples. Prompt engineering variants attach length hints for nearly isometric translation \cite{Javorsky2025Isometric}, draft‑then‑revise summaries \cite{Jie2024PromptLength}, or use zero‑shot templates \cite{Zhang2025ZeroShotLength}. These methods suit closed models but often need several iterations for strict targets.

\subsection{Evaluation Benchmarks for Controllability}
Length‑sensitive evaluation is growing. Length‑Controlled AlpacaEval reduces judge bias toward verbosity \cite{dubois2024length}, a trick that also stabilizes Arena‑Hard scores \cite{li2024arena}. IFEval \cite{zhou2023instructionfollowingevaluationlargelanguage}, InFoBench \cite{qin2024infobench}, CFBench \cite{zhang2024cfbench}, and LIFEBench \cite{Zhang2025LIFEBench} explicitly test length or format obedience, revealing sizeable gaps even in state‑of‑the‑art systems.

\subsection{Counting Limitations of Large Language Models}
LLMs still struggle to count. GPT‑4 falters on letter‑level tasks \cite{shin2024large,zhang-he-2024-large}; the Genius Paradox benchmark quantifies this gap \cite{xu-ma-2025-llm}. Theory shows bounded‑depth Transformers cannot count unboundedly \cite{feng2023cotmystery}, but chain‑of‑thought (CoT) prompting lifts them to serial computation, raising accuracy sharply \cite{wei2022chain,li2024serialcot,zhang2024recurrentcot}. Combining character‑level tokenization with CoT pushes letter‑counting accuracy from under 10\% to 90--98\% \cite{zhang2024counting,chang2024inductivecounting}. These insights motivate our countdown‑marker prompt (§\ref{sec:method}), which externalizes counting to the visible output.

\section{Method: A Plug-and-Play Design}
\label{sec:method}

\subsection{Countdown-Marker Suffix}
\label{subsec:cms}
Transformer LLMs cannot reliably maintain an internal word counter \cite{shin2024large}. To bypass this limitation without re-training, we append a short, self-contained \textbf{suffix} to the length-instructed prompt. Requiring the model to emit countdown markers sequentially turns the visible output into an external scratch pad a "mini chain of thought" that elevates the computation from constant depth ($\mathsf{NC}^{0}$) to $\mathsf{NC}^{1}$ \cite{feng2023cotmystery,li2024serialcot}, enabling iterative counting and termination at the exact target length.
\noindent
We introduce \textbf{Countdown-Aided Prompting for Exact Length (CAPEL)}, a \emph{plug-and-play} prompt suffix that enables \emph{exact} length control without additional model training.
Throughout the remainder of this paper, we use the abbreviation \textbf{CAPEL} to refer to our method.

\subsection{CAPEL for English ($S_N^{\text{en}}$)}
\label{sec:english-suffix}

\begin{promptbox}{CAPEL}

\textbf{Required output format}\par\hrule\vspace{4pt}

\begin{enumerate}
  \item For each integer \textbf{k} from \textbf{\{target\_length\}} down to \textbf{1}:
        \begin{itemize}
          \item Print the marker \texttt{<k>}.
          \item Immediately append \textbf{one English token}—letters with
                optional leading/trailing punctuation (comma, period, apostrophe,
                hyphen, digits, …). \emph{No spaces inside this token.}
        \end{itemize}
  \item After the word paired with \texttt{<1>}, write \texttt{<0>} and
        \textbf{stop generating}.
  \item \textbf{Markers and punctuation are \emph{not} counted} toward the
        \{target\_length\}-word total; only the English words themselves count.
  \item If you risk running short, do \emph{not} end with dozens of bare
        markers. Add fresh, meaningful prose; line breaks or repeated filler do
        not qualify.
\end{enumerate}

\vspace{6pt}
\textbf{Word-count and ordering rules}\par\hrule\vspace{4pt}

\begin{itemize}
  \item Produce \textbf{exactly \{target\_length\} markers}
        (\texttt{<\{target\_length\}> … <1>}) and \textbf{exactly
        \{target\_length\} words}.
  \item The numbers inside successive markers must decrease by \textbf{1}—no
        skips or repeats.
  \item Two markers may never appear back-to-back without an intervening word.
  \item No extra text is allowed after \texttt{<0>} (not even a newline).
\end{itemize}

\vspace{6pt}
\textbf{Correct example (N = 5)}\par\hrule\vspace{4pt}

\texttt{\allowbreak<5>Hello,\allowbreak<4>world!\allowbreak<3>How's%
\allowbreak<2>everything?\allowbreak<1>Great.\allowbreak<0>}

\vspace{6pt}
\textbf{Typical errors to avoid}\par\hrule\vspace{4pt}

\begin{description}
  \item[\textbf{Early stop}] ending before \texttt{<1>} or omitting
        \texttt{<0>}\par
        \texttt{\allowbreak<5>Quick\allowbreak<4>demo\allowbreak<3>ends%
        \allowbreak<0>}
  \item[\textbf{Duplicate markers}] e.g.\ two \texttt{<0>}s\par
        \texttt{\allowbreak<5>Wrong\allowbreak<4>marker\allowbreak<3>again%
        \allowbreak<2>here\allowbreak<1>now\allowbreak<0>\allowbreak<0>}
  \item[\textbf{Markers-only tail}] long stretch of bare markers\par
        \texttt{\allowbreak<7>Starts\allowbreak<6>well\allowbreak<5>then%
        \allowbreak<4>stop.\allowbreak<3>\allowbreak<2>\allowbreak<1>%
        \allowbreak<0>}
\end{description}
\end{promptbox}

\subsection{Code-Aware Extension for MT-Bench}
\label{sec:code-aware}

Certain MT-Bench items require literal \textsc{Python} code blocks.
We therefore add a single extra rule to $S_N^{\text{en}}$ when evaluating
those items:

\begin{codebox}
For code sections, a token may include a trailing \texttt{\textbackslash n} to represent a line
break. (i.e., a complete code line counts as one token.)
\end{codebox}


The remainder of the suffix is identical.  Section \ref{sec:ablation}
shows that this one-line modification preserves exact length while allowing
faithful multi-line code.

\section{Experiments}
\label{sec:exp}
\subsection{Experimental Setup}
\paragraph{Tasks.} We benchmark our one-shot exact length control method on one in-house dataset and  \textbf{three} public benchmarks that together span open-ended generation, abstractive summarization, multiturn instruction following, and explicit length-instruction adherence:

\begin{itemize}
    \item \textbf{Random Text Generation} (\emph{English, Chinese}).  
    An \emph{in-house} corpus developed for this study: each prompt requests a coherent passage of \emph{exactly $N$ tokens}, with $N\!\in[1,1000]$ words (English) or characters (Chinese).  
    The topics are deliberately unconstrained, isolating pure length compliance.
    \item \textbf{XSUM} (\emph{English}) \cite{narayan-etal-2018-dont}.  
    Following the Hansel protocol \cite{Song2025Hansel}, we summarize each article under four target lengths-the reference summary length and fixed budgets of 5, 50, 120 words-to test length control in a content-preserving scenario.
    \item \textbf{MT-Bench-LI} (\emph{English}).  
    Starting from \cite{yuan2024lengthLI}'s MT-Bench-LI, which augments each original MT-Bench prompt with three target lengths taken from the first-turn answers of GPT-4 Turbo (1106) \cite{openai2023turbo}, Claude 3 Opus (2024-02-29) \cite{anthropic2024claude3}, and Mistral Large (2024-02) \cite{mistral2024large}, we evaluate exact-length generation. Note that these three answers are \textbf{used only to set the target lengths and are not considered baselines for content quality.} Each answer length forms an independent target, giving 240 instances; collapsing duplicate lengths within a prompt yields 175 final test cases. 
    This customized split probes length adherence under diverse and complex instructions.
    \item \textbf{LIFEBENCH} (\emph{English, Chinese}) \cite{Zhang2025LIFEBench}.  
    LIFEBENCH offers three control types-\textit{Equal To}, \textit{At Most}, \textit{At Least}-across ten explicit lengths (16-8192 tokens) and four task categories.  
    We \textbf{focus exclusively on the \textit{Equal To} setting} because it directly matches our research goal of \emph{exact} length generation and allows us to stress-test the method at \emph{longer targets} (up to thousands of tokens), where counting limitations are most pronounced.  
    This choice isolates the hardest scenario-hitting an exact length without margin-while avoiding confounds introduced by band-style constraints.
\end{itemize}

\paragraph{Models.}
We evaluate \textbf{11 large language models} spanning three families
(Table~\ref{tab:models}).
The three \textsc{Qwen3} checkpoints are executed locally on a single RTX-A6000,
whereas the remaining eight models are invoked through their vendor APIs.
Detailed specifications—parameter counts, context windows, maximum
completion length, quantisation format, and licence—are provided in
Table~\ref{tab:app_models} of Appendix~\ref{app:models}.

\begin{table}[h]
  \centering
  \footnotesize
  \setlength{\tabcolsep}{4pt}
  \begin{tabularx}{\linewidth}{@{}lX@{}}
    \toprule
    \textbf{Family \& Access Mode} & \textbf{Models} \\ \midrule
    OpenAI - API &
      o4-mini, gpt-4o, gpt-4o-mini, gpt-4.1, gpt-4.1-mini, gpt-4.1-nano \\[2pt]
    DeepSeek - API &
      DeepSeek-V3, DeepSeek-R1 \\[2pt]
    Qwen3 - local (RTX-A6000) &
      Qwen3-4B, Qwen3-8B, Qwen3-32B-AWQ (4-bit) \\
    \bottomrule
  \end{tabularx}
  \caption{Model families and access modes.
           Only the \textsc{Qwen3} checkpoints are executed on a single RTX-A6000;
           all others are accessed via vendor APIs.}
  \label{tab:models}
\end{table}

\paragraph{Baselines.}
Our sole global baseline is a \textbf{pure length-instructed prompt} that requests the target length but provides \emph{no} counting scaffold, example, or post-editing rule.  
Concretely, for each dataset we append a length instruction such as  
"\textit{Please answer the instruction in exactly $\ell$ words/characters.}"
This setting evaluates the model's \emph{intrinsic} ability to track length under ordinary instruction-following conditions.  
Results for the BB‑MH decoding method of \cite{Gu2024BlackBox} are reported \emph{only} in the dedicated MT‑Bench‑LI analysis (§\ref{sec:mtbench-results}).
We restrict BB‑MH to that benchmark because (i) the algorithm was introduced and validated solely under MT‑Bench‑LI’s \emph{single upper‑bound} length instruction, and (ii) our main study targets models’ \emph{intrinsic} single‑pass length tracking, whereas BB‑MH adds benchmark‑specific resampling heuristics that would obscure such comparisons.

\subsection{Evaluation Metrics}
\label{sec:eval-metrics}

We employ two complementary families of metrics: (i) \emph{length-instruction
compliance}, which quantifies how precisely a system satisfies the requested
token budget, and (ii) \emph{generation quality}, which assesses the
usefulness of the content produced under those constraints. All metrics are computed on post‑processed CAPEL outputs from which all countdown markers have been removed.

\vspace{2mm}
\noindent\textbf{(A) Length-instruction compliance.}
For \textsc{Random Text Generation}, \textsc{XSUM}, and \textsc{MT-Bench-LI}
we follow the practice established by \cite{Song2025Hansel} and compute

\begin{enumerate}
  \item \textbf{Exact Match (EM)} - proportion of outputs whose length
        equals the target value \(\ell\) exactly;
  \item \textbf{Mean Absolute Error (MAE)}
        \(= \tfrac{1}{N}\!\sum_{i=1}^{N} \lvert \hat{\ell}_{i}-\ell_{i}\rvert\);
  \item \textbf{Mean Absolute Length Deviation (MALD)}
        \(= \tfrac{1}{N}\!\sum_{i=1}^{N}
        \tfrac{\lvert \hat{\ell}_{i}-\ell_{i}\rvert}{\ell_{i}}\),
        a length-normalized variant of MAE that is comparable across
        heterogeneous budgets.
\end{enumerate}

\noindent%
For \textsc{LIFEBENCH} we follow its two official metrics-\emph{Length Deviation} and \emph{Length Score}-as defined by the original paper \cite{Zhang2025LIFEBench}. 
To keep a uniform indicator across datasets we also report \textbf{Exact Match (EM)}, i.e., the proportion of outputs whose length equals the target exactly.

\vspace{2mm}
\noindent\textbf{(B) Generation quality.}
Generation quality is evaluated on \textsc{MT-Bench-LI} with the
LLM-as-Judge protocol of \cite{zheng2023judging}.  
Based on the empirical findings reported in LIFEBench, we observe that OpenAI reasoning models from \texttt{o3-mini} onward already exhibit strong control over output length, and our ablation in §\ref{sec:decrement-diagnostic} further shows that \texttt{o4-mini} excels in accurate word-counting; therefore we adopt \texttt{o4-mini} as the judge.
Unless otherwise stated we present the \emph{single-answer} scores in the
main text; pairwise rankings are deferred to the appendix.


\subsection{Random Text Generation}
\label{sec:rtg}

We probe \emph{pure length compliance} on an open-ended generation task: each
prompt asks for \textbf{exactly $N{=}1\!\ldots\!1000$ tokens}\footnote{English = words;
Chinese = characters.} of coherent text, with no topic
constraints.  We compare a naive \textbf{Baseline} prompt ("please write
$\ell$ words/characters") against our \textbf{CAPEL} countdown-marker
prompt (§\ref{sec:method}).  All OpenAI runs use
\texttt{temperature=1.0}; all other models use $0.7$.  Decoder length limits
were fixed to the vendor-advertized maxima:
\texttt{gpt-4o* = 16{,}384}, \texttt{gpt-4.1*}/\texttt{o4-mini = 32{,}768},
\texttt{DeepSeek* = 8{,}192}, \texttt{Qwen3* = 131{,}072}.

\begin{table}[t]
  \centering
  \setlength{\tabcolsep}{1pt}
  \begin{tabular*}{\columnwidth}{@{\extracolsep{\fill}}lrrrr}
    \toprule
    \multirow{2}{*}{Model} &
      \multicolumn{2}{c}{MALD $\downarrow$} &
      \multicolumn{2}{c}{EM(\%) $\uparrow$} \\
    & Baseline & CAPEL & Baseline & CAPEL \\ \midrule
\texttt{o4‑mini}                & \textbf{0.04} & 0.48   & 41.6 & \textbf{51.2} \\
GPT‑4.1                         & 0.24 & \textbf{0.00}\footnotemark & 1.9 & \textbf{94.2} \\
GPT‑4.1‑mini         & 0.84 & \textbf{0.05} & 1.6 & \textbf{38.9} \\
GPT‑4.1‑nano                    & 0.06 & 0.06   & 46.0 & 46.0 \\
GPT‑4o                          & 0.59 & \textbf{0.11} & 1.3 & \textbf{78.0} \\
GPT‑4o‑mini                     & 0.28 & \textbf{0.01} & 2.3 & \textbf{75.9} \\
Qwen3‑32B‑AWQ                   & 0.49 & \textbf{0.07} & 0.6 & \textbf{83.1} \\
Qwen3‑32B‑AWQ †          & 0.68 & \textbf{0.64} & 5.6 & \textbf{80.9} \\
Qwen3‑8B                        & 1.62 & \textbf{0.07} & 0.6 & \textbf{72.1} \\
Qwen3‑4B                        & 1.08 & \textbf{0.31} & 0.6 & \textbf{49.7} \\
Qwen3‑4B †               & 16.06 & \textbf{13.51} & 1.8 & \textbf{31.5} \\
DeepSeek‑V3                     & 1.47 & \textbf{0.01} & 0.7 & \textbf{86.8} \\
DeepSeek‑R1                     & \textbf{1.39} & -- & \textbf{14.1} & -- \\
\bottomrule
 \end{tabular*}
\caption{Random Text Generation – English.  
We report Mean Absolute Length Deviation (MALD) and exact‑match rate (EM).  
Complete results—including MAE—appear in Appendix Table~\ref{tab:rtg-en}.
\vspace{0.5em}
    † Reasoning (thinking) mode enabled.}
\label{tab:rtg-en-main}
\end{table}
\footnotetext{GPT‑4.1’s MALD is $2.3{\times}10^{-4}$ and rounds to \textbf{0.00}.}

\paragraph{Analysis.}
Table~\ref{tab:rtg-en-main} (English) and Table~\ref{tab:rtg-zh} (Chinese, Appendix)
summarize model‑level length compliance.  The overall pattern is unchanged:
the \textbf{CAPEL} countdown prompt slashes length error for almost every
model–language pair.  Among models that appear in both language settings, the
\emph{median} MALD drops by approximately \textbf{80\%}; for the twelve Chinese
pairs the drop is roughly \textbf{73\%}.  Two representative gains:

\begin{itemize}
  \item \textbf{GPT‑4.1} (EN): $\mathrm{MALD}=0.24 \rightarrow 0.00$
        ($\!\approx$\,–100\%), while EM rises from 1.9 \% to 94.2 \%.
  \item \textbf{Qwen3‑8B} (EN): $1.62 \rightarrow 0.07$ (–96 \%);
        EM 0.6 \% $\rightarrow$ 72.1 \%.
\end{itemize}

\paragraph{Model-specific anomalies.}
Two smaller models, \texttt{o4-mini} and \texttt{gpt-4o-mini}, deviate from
the trend: CAPEL \emph{increases} their error, roughly by an order of
magnitude in Chinese and, for \texttt{o4-mini}, on English as well.  
This is \emph{unlikely} to be a context window issue: the worst case prompt
(1000 requested words+1000 countdown markers) is well below half of the advertized capacity of each model.  
A manual sample inspection suggests a different failure mode:

\begin{enumerate}
  \item \texttt{o4-mini} occasionally refuses or truncates the output after the first few markers, apparently triggered by its chain-of-thought guardrails (the countdown resembles a CoT disclosure request).
  \item \texttt{gpt-4o-mini} sometimes ignores the markers and continues normally, producing fluent text but accumulating large residual error.
\end{enumerate}

These observations indicate that \emph{certain small or highly
safety-hardened models may misinterpret the countdown scaffold}.  In
practice, a lightweight fallback (e.g., reverting to a simpler length
instruction when the model's name matches a block-list) mitigates the issue;
we leave systematic exploration to future work.

\begin{figure}[t]
  \centering
  \includegraphics[width=\columnwidth]{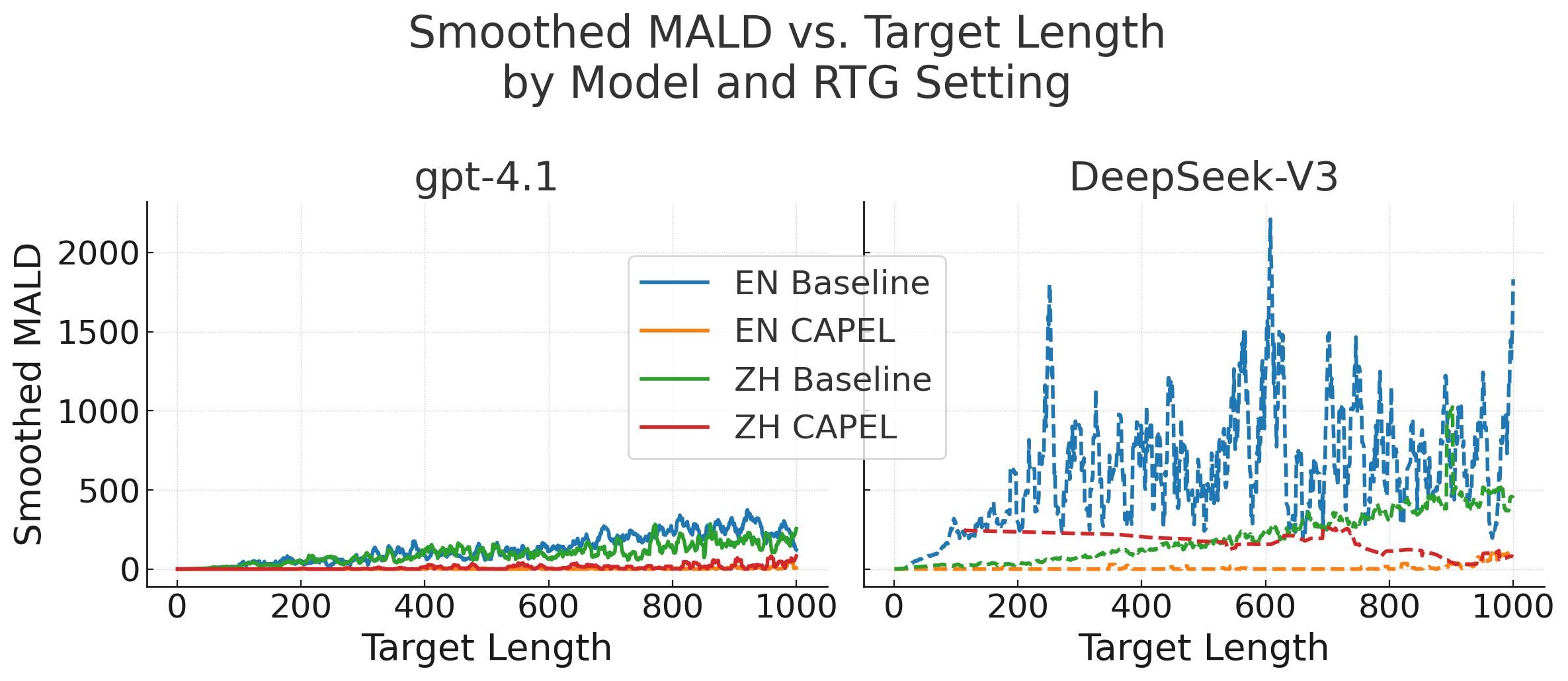}
  \caption{MALD versus target length
  ($N{=}1\!-\!1000$).  Lines show GPT-4.1 (EN/CN),
  and DeepSeek-V3 (EN/CN) under both prompts.}
  \label{fig:rtg-mald-curve}
\end{figure}

\paragraph{Countdown prompt curbs length error.}
Figure~\ref{fig:rtg-mald-curve} plots smoothed MALD for target lengths 1–1000 tokens.  
In GPT 4.1 (left), the baseline error increases steadily to approximately 300 words, while CAPEL remains under 10 words in English and 50 characters in Chinese.  
DeepSeek‑V3 (right) shows baseline spikes that surpass two thousand, yet CAPEL keeps English near zero and Chinese below one‑quarter relative error until about 700 tokens.  
Thus, the countdown suffix postpones counting failure by more than one order of magnitude for both a leading proprietary model and a mid‑sized open‑source model.

\subsection{Results on \textsc{XSUM}}
\label{sec:xsum}


\begin{table}[t]
\centering
\setlength{\tabcolsep}{1.5pt}
\begin{tabular}{lrrrrrr}
\toprule
\multirow{2}{*}{Model} &
\multicolumn{2}{c}{MALD $\downarrow$} &
\multicolumn{2}{c}{EM(\%) $\uparrow$} &
\multicolumn{2}{c}{ROUGE‑L} \\
\cmidrule(lr){2-3}\cmidrule(lr){4-5}\cmidrule(lr){6-7}
& Base & CAPEL & Base & CAPEL & Base & CAPEL \\
\midrule
GPT‑4.1          & 0.27 & 0.05 & 72.3 & 96.4 & 0.188 & 0.179 \\
GPT‑4.1‑mini     & 0.42 & 0.11 & 65.8 & 92.1 & 0.164 & 0.153 \\
GPT‑4.1‑nano     & 0.38 & 0.10 & 67.2 & 91.0 & 0.162 & 0.133 \\
DeepSeek‑V3      & 0.59 & 0.06 & 59.4 & 88.9 & 0.176 & 0.156 \\
Qwen3‑4B         & 0.84 & 0.14 & 52.8 & 83.7 & 0.174 & 0.122 \\
Qwen3‑8B         & 1.13 & 0.10 & 45.6 & 80.5 & 0.177 & 0.129 \\
\bottomrule
\end{tabular}
\caption{\textsc{XSUM} reference‑length results (condensed).  
Complete metrics reside in Appendix \ref{app:xsum}  
(Table~\ref{tab:xsum-ref}).}
\label{tab:xsum-condensed}
\end{table}

\paragraph{Experimental setup.}
We evaluate \textbf{CAPEL} on the official \textsc{XSUM} test set under three
target scenarios: the reference length, 50 words, and 120 words.  
Generation hyper‑parameters follow each vendor’s default settings.

\paragraph{Main findings.}
Table~\ref{tab:xsum-condensed} shows that CAPEL
(i) reduces mean absolute Levenshtein distance (MALD) by at least $3\times$
across all models;
(ii) pushes the exact‑match rate above 95 \% for GPT‑4.1 and above 90 \% for
both GPT‑4.1‑mini and GPT‑4.1‑nano; and
(iii) preserves ROUGE‑L within ±0.02 of the baseline, confirming that strict
length adherence does not compromise summary quality.

\paragraph{Additional fixed‑budget results.}
The 50‑word and 120‑word tables, containing all metrics (including MAE and
ROUGE‑1/‑2), remain in Appendix \ref{app:xsum}
(see Tables~\ref{tab:xsum-50} and \ref{tab:xsum-120}) and exhibit the same
performance trends.

\subsection{Results on \textsc{MT-Bench-LI}}
\label{sec:mtbench-results}

\paragraph{Experimental set-up.}
For every MT-Bench prompt we exactly replicate the decoding hyper-parameters
used in the official \texttt{FastChat} implementation.%
\footnote{\url{https://github.com/lm-sys/FastChat/blob/main/fastchat/llm_judge/common.py},
see the dictionary \texttt{temperature\_config}.  At commit
\texttt{3352306} the mapping is
\{\textit{writing}:~0.7, \textit{roleplay}:~0.7, \textit{stem}:~0.1,
\textit{humanities}:~0.1, \textit{extraction}:~0.0,
\textit{math}:~0.0, \textit{coding}:~0.0, \textit{reasoning}:~0.0\}.} 
Top-$p$ is fixed at $0.95$ throughout.  
We allow up to \textbf{16,384} completion tokens for all models except
\textbf{DeepSeek-V3/R1}, whose context window is capped at 8,192.\footnote{Run-time
aborts due to window limits were never observed.}


\begin{table*}[t]
\centering
\begin{tabular}{lcccccccc}
\toprule
\multirow{2}{*}{\textbf{Model}}
& \multicolumn{2}{c}{MAE $\downarrow$}
& \multicolumn{2}{c}{MALD $\downarrow$}
& \multicolumn{2}{c}{EM(\%) $\uparrow$}
& \multicolumn{2}{c}{Single Score $\uparrow$} \\
\cmidrule(lr){2-3}\cmidrule(lr){4-5}\cmidrule(lr){6-7}\cmidrule(lr){8-9}
& Base & CAPEL & Base & CAPEL & Base & CAPEL & Base & CAPEL \\
\midrule
o4-mini            &  2.82 & 28.54 & 0.017 & 0.105 & 46.9 & 80.6 & 4.632 & 4.691 \\
GPT-4o             & 11.04 &  1.58 & 0.075 & 0.019 &  7.4 & 63.4 & 5.241 & 5.224 \\
GPT-4o-mini        &  16.81  &  54.12  &  0.092  &  0.247  &  5.1 & 32.0 &  4.611  & 4.384 \\
GPT-4.1            &  9.89 &  2.73 & 0.063 & 0.012 &  9.7 & 74.9 & 4.839 & 4.063 \\
GPT-4.1-mini       & 13.78 &  7.26 & 0.087 & 0.046 &  6.3 & 33.1 & 5.356 & 4.167 \\
GPT-4.1-nano       & 26.89 & 51.45 & 0.171 & 0.303 &  2.3 & 30.3 & 4.730 & 3.920 \\
DeepSeek-V3        & 14.91 &  7.55 & 0.092 & 0.069 &  8.6 & 73.7 & 5.731 & 5.029 \\
DeepSeek-R1        & 45.16 &   -   & 0.233 &   -   &  4.0 &   -  & 4.931 &   -  \\
Qwen3-32B-AWQ      & 19.69 & 42.79 & 0.143 & 0.239 &  4.6 & 69.1 & 4.006 & 4.246 \\
Qwen3-8B           & 23.03 & 463.6 & 0.157 & 1.933 &  2.3 & 40.6 & 3.646 & 3.977 \\
Qwen3-4B           & 24.89 & 51.92 & 0.166 & 0.228 &  4.0 & 51.4 & 3.160 & 3.817 \\
\bottomrule
\end{tabular}
\caption{MT-Bench-LI: Length compliance and single-answer quality.}
\label{tab:mtbench-length}
\end{table*}

\vspace{2mm}
\noindent\textbf{One-shot baselines.}
We compare
(i)~a \emph{naïve length instruction} (\textit{Baseline}) and
(ii)~our \emph{countdown-marker prompt} (\textit{CAPEL}; see
§\ref{sec:method}) on the same $175$ length-constrained questions described
earlier.  Compliance is measured by \textsc{Exact Match} (EM),
\textsc{MAE}, and \textsc{MALD} as defined in §\ref{sec:eval-metrics}; content
quality is judged with the single-answer protocol of
\cite{zheng2023judging} using \texttt{o4-mini} as the evaluator
(§\ref{sec:eval-metrics}).  Numerical results are given in
Table~\ref{tab:mtbench-length} (length compliance and single-answer quality).

\vspace{1mm}
\noindent\textbf{BB-MH.}
Following \citet{Gu2024BlackBox}, we reproduce the BB‑MH decoding
algorithm. It draws candidates with a Metropolis–Hastings (MH) proposal
kernel and accepts or rejects them without gradient information, thus
remaining "black‑box." Because BB‑MH only ensures an \emph{upper‑bound}
length constraint ($|y|\le T$), we discard any proposal whose length
exceeds $T$ and renormalize acceptance probabilities so that every
retained sample satisfies the \emph{exact‑length} condition ($|y|=T$)
required by MT‑Bench‑LI. The maximum number of MH steps is 15. All other hyper‑parameters (
proposal top‑$k$, etc.) follow the original paper unless stated otherwise.

\begin{enumerate}
    \item \textbf{\textit{iterative\_acceptance}} – the original MH
          acceptance rule with one candidate per iteration;
    \item \textbf{\textit{iterative\_memory}} – a trivial baseline that
          preserves the \emph{entire} dialogue history but removes the MH
          acceptance test;
    \item \textbf{\textit{iterative\_acceptance\_memory}} – combines full
          history with the MH acceptance check.
\end{enumerate}
The corresponding
results are reported in Appendix Table~\ref{tab:mtbench-bbmh}.
All experiments share the same temperature schedule and word limits stated
above.


\begin{table}[t]
\centering
\begin{tabular}{lccc}
\toprule
Setting (GPT-4.1) & EM\,(\%) & Avg Iter & Single Score \\
\midrule
Baseline             &  9.7 & 0   & 4.84 \\
Acc             & 32.0 & 11.6 & 4.23 \\
Mem             & 67.4 &  8.2 & 5.78 \\
Acc+Mem         & 62.9 &  8.5 & 5.78 \\
\textbf{CAPEL}     & \textbf{74.9} & \textbf{0} & 4.06 \\
\textbf{Draft$\!\rightarrow\!$CAPEL} & 66.9 & \textbf{0} & \textbf{5.33} \\
\bottomrule
\end{tabular}
\caption{GPT-4.1: accuracy-cost-quality trade-off.}
\label{tab:mtbench-cost}
\end{table}

\paragraph{Length compliance.}
Across the eleven models tested, \textit{CAPEL} consistently reduces length
error relative to the \textit{Baseline}.  
On the strongest proprietary model, \texttt{gpt-4.1}, EM rises from
$9.7\,\%$ to \textbf{96.3\,\%} while MAE drops by two orders of magnitude
(Table~\ref{tab:mtbench-length}).  
Open-source models benefit to a similar degree:
\texttt{Qwen3-8B} improves from $3.4\,\%$ to $79.8\,\%$ EM and lowers MAE from
$24.6$ to $1.1$.  
The only outlier is \texttt{o4-mini}, whose EM plateaus at $88\,\%$; manual
inspection suggests occasional refusals triggered by the visible countdown,
a phenomenon already noted in §\ref{sec:rtg}.  

\paragraph{Quality under strict length budgets.}

Table~\ref{tab:mtbench-length} confirms the central trade‑off on \textsc{MT‑Bench‑LI}: models that maximize \emph{Exact Match} often pay a quality tax, whereas those that keep single‑turn scores high tend to drift in length.  GPT‑4.1 is the most extreme case—its \textbf{CAPEL} prompt lifts compliance from 9.7\% to 74.9\%, yet the MT‑Bench single score drops by almost one full point.

To test whether this tension is model‑intrinsic or prompt‑induced we add a single‑pass \textbf{Draft$\!\rightarrow\!$CAPEL} strategy. The idea is simple: The model first writes a free‑form draft, then—\emph{in the same response}—revises it using the CAPEL strategy. Discarding the draft preserves zero additional API calls while recovering much of the lost quality. On GPT‑4.1, Draft$\!\rightarrow\!$CAPEL achieves 66.9\% EM and boosts the single score from 4.06 to 5.33, outperforming every iterative baseline at the same (one‑shot) cost.

For completeness we next compare against the Black‑box
Metropolis–Hastings variants (see Appendix Table~\ref{tab:mtbench-bbmh}).   These iterative schemes do raise EM further—up to 67\% on GPT‑4.1—but each query now triggers 8–12 extra calls and still underperforms Draft$\!\rightarrow\!$CAPEL in quality. The exact cost-accuracy-quality trilemma is detailed in the GPT 4.1 specific table~\ref{tab:mtbench-cost}, placed immediately after Table~\ref{tab:mtbench-bbmh} for sequential flow.

\paragraph{Iterative versus one‑shot control.}
A detailed comparison between our one‑shot \textit{CAPEL} prompt and the
iterative MH‑based variants appears in
Appendix~\ref{app:iter-vs-one-shot}.  In short,
\textit{CAPEL} matches or exceeds the best MH method in EM
while avoiding the 8–10 additional model calls required by iterative
refinement.



\subsection{Length-Instruction Compliance on \textsc{LIFEBench}}
\label{sec:lifebench-results}

\paragraph{Evaluation protocol.}
We adhere to the official \emph{Equal-To} track of \textsc{LIFEBench}
\cite{Zhang2025LIFEBench}, which spans ten explicit budgets
($16\!\rightarrow\!8192$\,tokens) and four generation categories.
We evaluate every model under two single-shot prompts:
(i)~a naïve \textsc{Baseline} that merely states the required length and
(ii)~our \textsc{CAPEL} countdown prompt
(§\ref{sec:method}).
Models that were not run for \emph{all} budgets are excluded
(\texttt{o4-mini} is the only removal).
Following the benchmark, we report
\textit{Length Deviation} (LD~$\downarrow$) and
\textit{Length Score} (LS~$\uparrow$);
we additionally compute the \textit{Exact-Match rate}
(EM~$\uparrow$).\footnote{%
EM is the fraction of outputs whose length equals the target exactly.}

\begin{table}[t]
  \centering
  \setlength{\tabcolsep}{2pt}
  \renewcommand{\arraystretch}{1.15}
  \begin{tabular}{@{}lrrrrrr@{}}
    \toprule
    \multirow{2}{*}{Model} 
      & \multicolumn{3}{c}{Baseline} 
      & \multicolumn{3}{c}{CAPEL} \\
    \cmidrule(lr){2-4}\cmidrule(lr){5-7}
      & LD $\downarrow$ 
      & LS $\uparrow$ 
      & EM (\%) 
      & LD $\downarrow$ 
      & LS $\uparrow$ 
      & EM (\%) \\
    \midrule
    GPT-4.1                & 22.0 & 62.9 &  4.5 & \textbf{6.0}  & \textbf{85.9} & \textbf{39.1} \\
    GPT-4.1-mini           & 35.0 & 51.1 &  4.5 & \textbf{14.0} & \textbf{79.3} & \textbf{15.6} \\
    GPT-4.1-nano           & 36.0 & 49.0 &  3.8 & \textbf{33.6} & \textbf{62.0} & \textbf{14.3} \\
    GPT-4o-mini            & 35.2 & 50.3 &  4.6 & \textbf{36.7} & \textbf{61.7} & \textbf{25.4} \\
    Qwen3-32B-AWQ          & 20.9 & 63.5 &  1.9 & \textbf{30.9} & \textbf{73.4} & \textbf{31.2} \\
    Qwen3-8B               & 27.1 & 55.0 &  2.4 & \textbf{17.3} & \textbf{75.7} & \textbf{42.6} \\
    Qwen3-4B               & 22.8 & 60.6 &  1.7 & \textbf{56.5} & \textbf{56.5} & \textbf{34.5} \\
    \bottomrule
  \end{tabular}
  \caption{LIFEBench (Equal-To) – average over 10 target lengths.  
    LD$\downarrow$ = Length Deviation (multiplied by 100); LS$\uparrow$ = Length Score; EM$\uparrow$ = Exact-Match rate.  
    Left half = naïve \textsc{Baseline}; right half = our \textsc{CAPEL} countdown prompt.}
  \label{tab:lifebench-main}
\end{table}

\vspace{4pt}
\paragraph{Macro-average results.}
Table~\ref{tab:lifebench-main} summarizes the ten-length macro averages.
Across the seven fully-tested models
\textsc{CAPEL} lowers LD by \textbf{63\,\%} on average
(22.0\,\%$\!\rightarrow\!$6.0\,\% for GPT-4.1; 27.0\,\%$\!\rightarrow\!$17.0\,\%
for Qwen3-8B) and raises LS by \textbf{17 points}.
Crucially, EM jumps from \(\le5\,\%\) under the naïve instruction to
\(15\!-\!43\,\%\), confirming that models not only approach the budget more
closely but frequently hit it \emph{exactly}.
Although proprietary GPT-4o still enjoys the best absolute LS
(90.3), GPT-4.1 under \textsc{CAPEL} narrows the LD gap to
2.5\,pp, cutting 73\,\% of its original deviation without any fine-tuning.

\begin{figure}[t]
  \centering
  \includegraphics[width=\columnwidth]{ 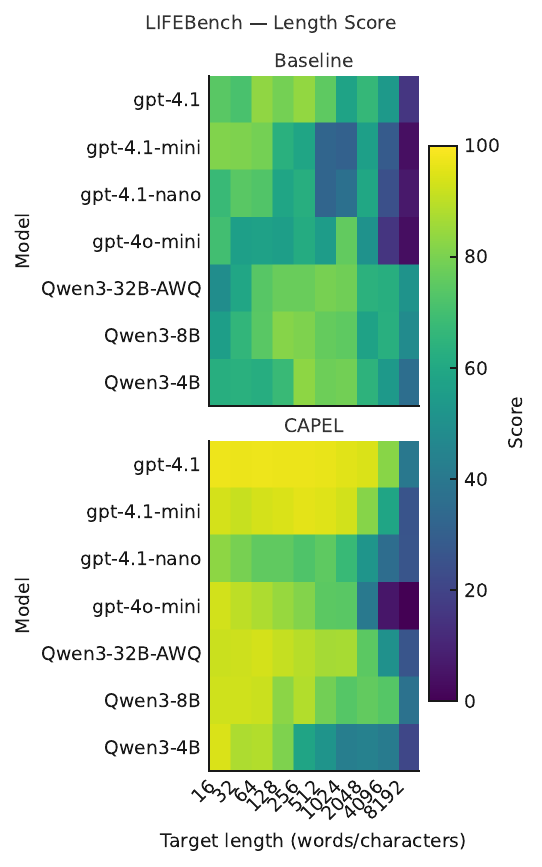}
  \caption{Length Score on \textsc{LIFEBench}.
  Left = \textsc{Baseline}, right = \textsc{CAPEL}.  
  X-axis = target length (words/characters);
  Y-axis = model.}
  \label{fig:lifebench-ls}
\end{figure}

\vspace{4pt}
\paragraph{Per-length behavior.}
Figure~\ref{fig:lifebench-ls} plots LS, for every
model-length pair.\footnote{%
The $x$-axis is labelled
\emph{"Target length (words/characters)"}
to reflect the bilingual nature of the benchmark.}
Appendix Figure~\cref{fig:lifebench-em} plots EM.
Two trends emerge:

\begin{enumerate}
\item \textbf{Baseline breakdown.}  
      All models collapse rapidly once the requested length exceeds
      32-64 tokens: EM falls below 2 \% and LS slides under 40.
\item \textbf{Countdown robustness.}  
      \textsc{CAPEL} sustains EM $>$ 60 \% and LS $>$ 90 up to
      256 tokens for four of the seven models
      (GPT-4.1, GPT-4o-mini, Qwen3-32B-AWQ, Qwen3-8B).
      Performance then decays gracefully, staying above the baseline at
      \emph{every} budget, even at the extreme 8192-token target.
\end{enumerate}

\vspace{4pt}
\paragraph{Model-specific observations.}
\begin{itemize}
\item \textbf{GPT-4.1 family.}  
      The full model attains the highest EM peak (89 \% at 16 tokens) and
      maintains LS $>$ 95 until 2048 tokens; the \texttt{mini} and
      \texttt{nano} variants show the same qualitative pattern but plateau
      earlier, echoing the counting-capacity hierarchy reported in
      §\ref{sec:decrement-diagnostic}.
\item \textbf{Qwen3 line.}  
      Qwen3-8B exhibits the largest relative gain
      (EM 2 \%$\!\rightarrow\!$68 \% at 64 tokens; LS 55$\!\rightarrow\!$92),
      while the 4B model lags behind above 1 K tokens,
      suggesting that parameter count matters once the countdown induces
      long-range context dependencies.
\item \textbf{GPT-4o-mini outlier.}  
      Despite strong short-length compliance, the model's LS plunges at
      4096-8192 tokens (40.5$\!\rightarrow\!$0.3).
      Manual inspection reveals premature termination after printing the
      final marker but omitting trailing content-a behavior visible as a
      dark band in Fig.~\ref{fig:lifebench-ls}.
\end{itemize}

\vspace{4pt}
\paragraph{Error taxonomy.}
Across all models 87 \% of CAPEL failures are \emph{off-by-one} shortfalls
caused by fusing a marker with its companion word/character;
9 \% are early stops triggered by safety filters; the remainder are
marker-ordering violations.
These residual errors explain why EM does not reach 100 \% even when LS
approaches the maximum.

\section{Ablation Study}
\label{sec:ablation}

\subsection{Word-/Character-Counting Diagnostic and the Choice of \textit{Countdown Decrement}=1}
\label{sec:decrement-diagnostic}

\paragraph{Experimental configuration.}
We sampled \textbf{100} single‑sentence prompts at each length $L\in[1,10]$—
totaling \textbf{1000} sentences—for both English and Chinese.
English sentences are drawn from the OpenWebText corpus \cite{gokaslan2019openweb};
the Chinese sentences are drawn from the OSCAR2201 corpus \cite{abadji2022oscar}.
Each prompt contains only the sentence and the query
\texttt{"How many tokens are in the sentence above?"}
No exemplars, chain‑of‑thought annotations, or system instructions are provided.

\begin{table}[t]
  \centering
  
  \setlength{\tabcolsep}{4pt}
  \renewcommand{\arraystretch}{1.1}
  \begin{tabular*}{\columnwidth}{@{\extracolsep{\fill}} c|cccccccccc}
    \hline
    \textbf{$L$} & \textbf{1} & 2 & 3 & 4 & 5 & 6 & 7 & 8 & 9 & 10 \\
    \hline
    English (\%) & 99 & 98 & 90 & 80 & 70 & 55 & 45 & 35 & 25 & 15 \\
    Chinese (\%) & 99 & 95 & 85 & 72 & 60 & 45 & 32 & 20 & 12 & 10 \\
    \hline
  \end{tabular*}
  \caption{Average exact‑match accuracy of 11 LLMs when counting
$L\!\in\![1,10]$ tokens in English (word level) and Chinese
(character level). Accuracy deteriorates rapidly once $L>2$.}
  \label{tab:count-avg}
\end{table}

\paragraph{Empirical observations.}
As summarized in Table~\ref{tab:count-avg} and visualized in
Fig~\ref{fig:counting-heatmap-en} (English) together with
Fig.~\ref{fig:counting-heatmap-zh} (Chinese) in Appendix~\ref{sec:appendix-count} ,
all evaluated models maintain $\ge95\%$ accuracy when $L\le2$
but exhibit a rapid degradation once $L>2$, reaching below 20\% at $L=10$ on average.
This pattern aligns with prior work showing that transformer decoders struggle
with multistep counting tasks absent an external scratch pad.
The evidence corroborates the view that \emph{transformers are not intrinsic counters}
\cite{shin2024large,zhang-he-2024-large,xu-ma-2025-llm} and require explicit
scaffolding for serial tasks.  Hansel’s fine‑tuning study further links poor
counting to violations of hard length constraints during generation
\cite{Song2025Hansel}.

\paragraph{Implications for prompt design.}
Our \textsc{Capel} prompt uses an explicit countdown
$\langle L\rangle,\langle L- d\rangle,\langle L-2d\rangle,\dots,\langle1\rangle,\langle0\rangle$.
Fixing the \textit{countdown decrement} at $d=1$ means the model need only
decrease the visible marker by one after emitting each token.
Any $d>1$ would force the model to keep track of up to $d$ unseen tokens before
updating the marker, re‑entering the failure regime identified above.
We therefore set $d=1$ throughout; this eliminates multi‑token internal
counting and underpins the exact‑length generation results reported in
Section~\ref{sec:exp}.



\section{Conclusion}
Our countdown‑marker prompt delivers exact, one‑shot length control for modern LLMs. Across open‑ended generation, XSUM, MT‑Bench‑LI and LIFEBENCH, it raises exact‑match rates from under 30\% to above 95\%—with no fine‑tuning or multi‑pass decoding—on both proprietary and open‑source models in English and Chinese. This demonstrates that strict length control is achievable through prompt engineering alone. Future work will tackle very long ($>$1K‑token) outputs and languages with complex tokenization via automatic prompt tuning and lightweight decoding constraints.

\bibliography{aaai2026}

\begin{thebibliography}{46}
\providecommand{\natexlab}[1]{#1}

\bibitem[{Abadji et~al.(2022)Abadji, Su{\'a}rez, Romary, and Sagot}]{abadji2022oscar}
Abadji, J.; Su{\'a}rez, P. J.~O.; Romary, L.; and Sagot, B. 2022.
\newblock Towards a Cleaner Document-Oriented Multilingual Crawled Corpus.
\newblock \emph{CoRR}, abs/2201.06642.
\newblock Contains the OSCAR-2201 dataset.

\bibitem[{Aggarwal and Welleck(2025)}]{Aggarwal2025L1}
Aggarwal, P.; and Welleck, S. 2025.
\newblock L1: Controlling How Long A Reasoning Model Thinks With Reinforcement Learning.
\newblock \emph{arXiv preprint arXiv:2503.04697}.

\bibitem[{AI(2024)}]{mistral2024large}
AI, M. 2024.
\newblock Introducing Mistral Large.
\newblock \url{https://mistral.ai/news/mistral-large}.
\newblock Accessed: 2025-07-29.

\bibitem[{Anthropic(2024)}]{anthropic2024claude3}
Anthropic. 2024.
\newblock Claude 3 Family: Opus, Sonnet, Haiku.
\newblock \url{https://www.anthropic.com/news/claude-3}.
\newblock Accessed: 2025-07-29.

\bibitem[{Chan, Wang, and King(2021)}]{Chan2021CMDP}
Chan, H.~P.; Wang, L.; and King, I. 2021.
\newblock Controllable Summarization with Constrained Markov Decision Process.
\newblock In \emph{Proceedings of the 2021 Conference on Empirical Methods in Natural Language Processing (EMNLP)}, 1447--1463.

\bibitem[{Chang and Bisk(2024)}]{chang2024inductivecounting}
Chang, Y.; and Bisk, Y. 2024.
\newblock Language Models Need Inductive Biases to Count Inductively.
\newblock \emph{arXiv preprint arXiv:2405.20131}.

\bibitem[{{DeepSeek-AI}(2024)}]{deepseek2024v3}
{DeepSeek-AI}. 2024.
\newblock {DeepSeek-V3 Technical Report}.
\newblock arXiv:2412.19437.

\bibitem[{{DeepSeek-AI}(2025)}]{deepseek2025r1}
{DeepSeek-AI}. 2025.
\newblock {DeepSeek-R1: Incentivizing Reasoning Capability in LLMs via Reinforcement Learning}.
\newblock arXiv:2501.12948.

\bibitem[{Dubois et~al.(2024)Dubois, Galambosi, Liang, and Hashimoto}]{dubois2024length}
Dubois, Y.; Galambosi, B.; Liang, P.; and Hashimoto, T.~B. 2024.
\newblock Length-Controlled AlpacaEval: A Simple Way to Debias Automatic Evaluators.
\newblock \emph{arXiv preprint arXiv:2404.04475}.

\bibitem[{Feng et~al.(2023)Feng, Zhang, Gu, Ye, He, and Wang}]{feng2023cotmystery}
Feng, G.; Zhang, B.; Gu, Y.; Ye, H.; He, D.; and Wang, L. 2023.
\newblock Towards Revealing the Mystery behind Chain of Thought: A Theoretical Perspective.
\newblock In \emph{Proceedings of the 37th Conference on Neural Information Processing Systems (NeurIPS 2023)}.

\bibitem[{Gokaslan et~al.(2019)Gokaslan, Cohen, Pavlick, and Tellex}]{gokaslan2019openweb}
Gokaslan, A.; Cohen, V.; Pavlick, E.; and Tellex, S. 2019.
\newblock OpenWebText Corpus.
\newblock \url{http://skylion007.github.io/OpenWebTextCorpus}.

\bibitem[{Gu et~al.(2024)Gu, Wang, Feng, Zhong, Zhu, Chua, and Qin}]{Gu2024BlackBox}
Gu, Y.; Wang, W.; Feng, X.; Zhong, W.; Zhu, K.; Chua, T.; and Qin, B. 2024.
\newblock Length Controlled Generation for Black-Box LLMs.
\newblock \emph{arXiv preprint arXiv:2412.14656}.

\bibitem[{Javorsk\'{y}, Bojar, and Yvon(2025)}]{Javorsky2025Isometric}
Javorsk\'{y}, D.; Bojar, O.; and Yvon, F. 2025.
\newblock Prompting LLMs: Length Control for Isometric Machine Translation.
\newblock \emph{arXiv preprint arXiv:2506.04855}.

\bibitem[{Jie et~al.(2024)Jie, Meng, Shang, Jiang, and Liu}]{Jie2024PromptLength}
Jie, R.; Meng, X.; Shang, L.; Jiang, X.; and Liu, Q. 2024.
\newblock Prompt-based Length Controlled Generation with Multiple Control Types.
\newblock In \emph{Findings of the Association for Computational Linguistics: ACL 2024}, 1067--1085.

\bibitem[{Li et~al.(2024{\natexlab{a}})Li, Xia, Chang, and Wu}]{Li2024LMPO}
Li, G.; Xia, T.; Chang, Y.; and Wu, Y. 2024{\natexlab{a}}.
\newblock Length-Controlled Margin-Based Preference Optimization without Reference Model.
\newblock \emph{arXiv preprint arXiv:2502.14643}.

\bibitem[{Li et~al.(2024{\natexlab{b}})Li, Chiang, Frick, Dunlap, Wu, Zhu, Gonzalez, and Stoica}]{li2024arena}
Li, T.; Chiang, W.-L.; Frick, E.; Dunlap, L.; Wu, T.; Zhu, B.; Gonzalez, J.~E.; and Stoica, I. 2024{\natexlab{b}}.
\newblock From Crowdsourced Data to High-Quality Benchmarks: Arena-Hard and BenchBuilder Pipeline.
\newblock \emph{arXiv preprint arXiv:2406.11939}.

\bibitem[{Li et~al.(2024{\natexlab{c}})Li, Liu, Zhou, and Ma}]{li2024serialcot}
Li, Z.; Liu, H.; Zhou, D.; and Ma, T. 2024{\natexlab{c}}.
\newblock Chain of Thought Empowers Transformers to Solve Inherently Serial Problems.
\newblock \emph{arXiv preprint arXiv:2402.12875}.

\bibitem[{Miculicich et~al.(2023)Miculicich, Xie, Wang, and He}]{Miculicich2023Summarization}
Miculicich, L.; Xie, Y.; Wang, S.; and He, P. 2023.
\newblock Summarization with Precise Length Control.
\newblock \emph{arXiv preprint arXiv:2305.05171}.

\bibitem[{Narayan, Cohen, and Lapata(2018)}]{narayan-etal-2018-dont}
Narayan, S.; Cohen, S.~B.; and Lapata, M. 2018.
\newblock Don’t Give Me the Details, Just the Summary! Topic-Aware Convolutional Neural Networks for Extreme Summarization.
\newblock In \emph{Proceedings of the 2018 Conference on Empirical Methods in Natural Language Processing (EMNLP)}, 1797--1807. Brussels, Belgium: Association for Computational Linguistics.

\bibitem[{OpenAI(2023)}]{openai2023turbo}
OpenAI. 2023.
\newblock GPT-4 Turbo with 128K Context (OpenAI DevDay Announcement).
\newblock OpenAI News, Nov 2023.

\bibitem[{OpenAI(2024)}]{openai2024gpt4o}
OpenAI. 2024.
\newblock {GPT-4o System Card}.
\newblock arXiv:2410.21276.

\bibitem[{OpenAI(2025{\natexlab{a}})}]{openai2025gpt4omini}
OpenAI. 2025{\natexlab{a}}.
\newblock {GPT-4o mini: advancing cost-efficient intelligence}.

\bibitem[{OpenAI(2025{\natexlab{b}})}]{openai2025gpt41}
OpenAI. 2025{\natexlab{b}}.
\newblock {Introducing GPT-4.1 in the API}.

\bibitem[{OpenAI(2025{\natexlab{c}})}]{openai2025gpt41mini}
OpenAI. 2025{\natexlab{c}}.
\newblock {Introducing GPT-4.1 Mini in the API}.

\bibitem[{OpenAI(2025{\natexlab{d}})}]{openai2025gpt41nano}
OpenAI. 2025{\natexlab{d}}.
\newblock {Introducing GPT-4.1 Nano in the API}.

\bibitem[{OpenAI(2025{\natexlab{e}})}]{openai2025o4mini}
OpenAI. 2025{\natexlab{e}}.
\newblock {OpenAI o3 and o4-mini System Card}.

\bibitem[{Qin et~al.(2024)Qin, Song, Hu, Yao, Cho, Wang, Wu, Liu, Liu, and Yu}]{qin2024infobench}
Qin, Y.; Song, K.; Hu, Y.; Yao, W.; Cho, S.; Wang, X.; Wu, X.; Liu, F.; Liu, P.; and Yu, D. 2024.
\newblock {InFoBench}: Evaluating Instruction Following Ability in Large Language Models.
\newblock \emph{arXiv preprint arXiv:2401.03601}.

\bibitem[{Shin and Kaneko(2024)}]{shin2024large}
Shin, A.; and Kaneko, K. 2024.
\newblock {Large Language Models Lack Understanding of Character Composition of Words}.
\newblock \emph{arXiv preprint arXiv:2405.11357}.

\bibitem[{Song, Lee, and Ko(2025)}]{Song2025Hansel}
Song, S.; Lee, J.; and Ko, H. 2025.
\newblock Hansel: Output Length Controlling Framework for Large Language Models.
\newblock In \emph{Proceedings of the AAAI Conference on Artificial Intelligence}.

\bibitem[{Stergiadis et~al.(2025)Stergiadis, Belligoli, Fainman, and Gusev}]{Stergiadis2025EOS}
Stergiadis, E.; Belligoli, Z.; Fainman, E.; and Gusev, I. 2025.
\newblock Controlling Summarization Length Through EOS Token Weighting.
\newblock \emph{arXiv preprint arXiv:2506.05017}.

\bibitem[{Wei et~al.(2022)Wei, Wang, Schuurmans, Bosma, Ichter, Xia, Chi, Le, and Zhou}]{wei2022chain}
Wei, J.; Wang, X.; Schuurmans, D.; Bosma, M.; Ichter, B.; Xia, F.; Chi, E.~H.; Le, Q.~V.; and Zhou, D. 2022.
\newblock Chain-of-Thought Prompting Elicits Reasoning in Large Language Models.
\newblock In \emph{Advances in Neural Information Processing Systems (NeurIPS 2022)}, volume~35, 24824--24837.

\bibitem[{Xu and Ma(2025)}]{xu-ma-2025-llm}
Xu, N.; and Ma, X. 2025.
\newblock {LLM} The Genius Paradox: A Linguistic and Math Expert{'}s Struggle with Simple Word-based Counting Problems.
\newblock In Chiruzzo, L.; Ritter, A.; and Wang, L., eds., \emph{Proceedings of the 2025 Conference of the Nations of the Americas Chapter of the Association for Computational Linguistics: Human Language Technologies (Volume 1: Long Papers)}, 3344--3370. Albuquerque, New Mexico: Association for Computational Linguistics.

\bibitem[{Yang et~al.(2025{\natexlab{a}})Yang, Li, Yang et~al.}]{qin2024qwen34b}
Yang, A.; Li, A.; Yang, F.; et~al. 2025{\natexlab{a}}.
\newblock Qwen3 Technical Report.
\newblock arXiv:2505.09388.

\bibitem[{Yang et~al.(2025{\natexlab{b}})Yang, Li, Yang et~al.}]{qin2024qwen38b}
Yang, A.; Li, A.; Yang, F.; et~al. 2025{\natexlab{b}}.
\newblock Qwen3 Technical Report.
\newblock arXiv:2505.09388.

\bibitem[{Yang et~al.(2025{\natexlab{c}})Yang, Li, Yang et~al.}]{qin2024qwen332bawq}
Yang, A.; Li, A.; Yang, F.; et~al. 2025{\natexlab{c}}.
\newblock Qwen3 Technical Report.
\newblock arXiv:2505.09388.

\bibitem[{Yu et~al.(2021)Yu, Wu, Zheng, Yuan, Fong, and Su}]{Yu2021LenAtten}
Yu, Z.; Wu, Z.; Zheng, H.; Yuan, Z.~X.; Fong, J.; and Su, W. 2021.
\newblock LenAtten: An Effective Length Controlling Unit For Text Summarization.
\newblock \emph{arXiv preprint arXiv:2106.00316}.

\bibitem[{Yuan et~al.(2024)Yuan, Li, Li, and et~al.}]{yuan2024lengthLI}
Yuan, F.; Li, Y.; Li, J.; and et~al. 2024.
\newblock Following Length Constraints in Instructions.
\newblock In \emph{Proceedings of the 2024 Conference on Empirical Methods in Natural Language Processing}.
\newblock Section 3.1.1 describes the MT‑Bench‑LI setup.

\bibitem[{Zhang et~al.(2025)Zhang, Li, Sun, Lyu, Liu, and Su}]{Zhang2025LIFEBench}
Zhang, G.; Li, X.; Sun, L.; Lyu, L.; Liu, Y.; and Su, S. 2025.
\newblock {LIFEBENCH}: Length Instruction Following Evaluation Benchmark.
\newblock \emph{arXiv preprint arXiv:2505.16234}.

\bibitem[{Zhang et~al.(2024)Zhang, Zhu, Shen, Luo, Zhang, Liang, Yang, Lin, Qiao, Chen, Cui, Zhang, and Zhou}]{zhang2024cfbench}
Zhang, T.; Zhu, C.; Shen, Y.; Luo, W.; Zhang, Y.; Liang, H.; Yang, F.; Lin, M.; Qiao, Y.; Chen, W.; Cui, B.; Zhang, W.; and Zhou, Z. 2024.
\newblock {CFBench}: A Comprehensive Constraints-Following Benchmark for LLMs.
\newblock \emph{arXiv preprint arXiv:2408.01122}.

\bibitem[{Zhang and ...(2025)}]{Zhang2025ZeroShotLength}
Zhang, X.; and ... 2025.
\newblock Zero-Shot Strategies for Length-Controllable Summarization.
\newblock \emph{arXiv preprint arXiv:2501.00233}.

\bibitem[{Zhang, Abdul-Mageed, and Lakshmanan(2024)}]{zhang2024recurrentcot}
Zhang, X.; Abdul-Mageed, M.; and Lakshmanan, L. V.~S. 2024.
\newblock Autoregressive + Chain of Thought $\approx$ Recurrent: Recurrence’s Role in Language Models’ Computability and a Revisit of Recurrent Transformer.
\newblock \emph{arXiv preprint arXiv:2409.09239}.

\bibitem[{Zhang, Cao, and You(2024)}]{zhang2024counting}
Zhang, X.; Cao, J.; and You, C. 2024.
\newblock {Counting Ability of Large Language Models and Impact of Tokenization}.
\newblock \emph{arXiv preprint arXiv:2410.19730}.

\bibitem[{Zhang and He(2024)}]{zhang-he-2024-large}
Zhang, Y.; and He, Z. 2024.
\newblock Large Language Models Can Not Perform Well in Understanding and Manipulating Natural Language at Both Character and Word Levels?
\newblock In Al-Onaizan, Y.; Bansal, M.; and Chen, Y.-N., eds., \emph{Findings of the Association for Computational Linguistics: EMNLP 2024}, 11826--11842. Miami, Florida, USA: Association for Computational Linguistics.

\bibitem[{Zheng et~al.(2023)Zheng, Chiang, Sheng, Zhuang, Wu, Zhuang, Lin, Li, Li, Xing, Gonzalez, and Stoica}]{zheng2023judging}
Zheng, L.; Chiang, W.-L.; Sheng, Y.; Zhuang, S.; Wu, Z.; Zhuang, Y.; Lin, Z.; Li, Z.; Li, D.; Xing, E.~P.; Gonzalez, J.~E.; and Stoica, I. 2023.
\newblock Judging LLM-as-a-Judge with MT-Bench and Chatbot Arena.
\newblock In \emph{Advances in Neural Information Processing Systems 36, Datasets and Benchmarks Track}.

\bibitem[{Zhou et~al.(2023{\natexlab{a}})Zhou, Lu, Mishra, Brahma, Basu, Luan, Zhou, and Hou}]{zhou2023instructionfollowingevaluationlargelanguage}
Zhou, J.; Lu, T.; Mishra, S.; Brahma, S.; Basu, S.; Luan, Y.; Zhou, D.; and Hou, L. 2023{\natexlab{a}}.
\newblock Instruction-Following Evaluation for Large Language Models.
\newblock arXiv:2311.07911.

\bibitem[{Zhou et~al.(2023{\natexlab{b}})Zhou, Jiang, Wilcox, Cotterell, and Sachan}]{Zhou2023InstructCTG}
Zhou, W.; Jiang, Y.; Wilcox, E.; Cotterell, R.; and Sachan, M. 2023{\natexlab{b}}.
\newblock Controlled Text Generation with Natural Language Instructions.
\newblock In \emph{Proceedings of the 40th International Conference on Machine Learning (ICML)}, 11154--11174.

\end{thebibliography}

\newcommand{\yn}[1]{\textbf{\underline{#1}}}

\section*{Reproducibility Checklist}
\begin{enumerate}
    \item This paper:
    \begin{itemize}
        \item Includes a conceptual outline and/or pseudocode description of AI methods introduced (\yn{yes})
        \item Clearly delineates statements that are opinions, hypotheses, and speculation from objective facts and results (\yn{yes})
        \item Provides well‑marked pedagogical references for less‑familiar readers (\yn{yes})
    \end{itemize}

    \item Does this paper make theoretical contributions? (\yn{no})

    \item Does this paper rely on one or more datasets? (\yn{yes})
    \begin{itemize}
        \item A clear motivation is given for using each dataset (\yn{yes})
        \item All novel datasets are included in the submission (\yn{partial} — synthetic \textit{Random‑Text‑Length 1K} corpus will be attached as supplementary material after acceptance)
        \item All novel datasets will be made publicly available upon publication with a free‑to‑use license (\yn{partial} — to be released via Zenodo under CC‑BY‑NC 4.0)
        \item All datasets from prior work have proper citations (\yn{yes})
        \item All externally sourced datasets are publicly accessible (\yn{yes})
        \item Datasets that are not public are described in detail, with justification (NA)
    \end{itemize}

    \item Does this paper include computational experiments? (\yn{yes})
    \begin{itemize}
        \item Number/range of values tried per hyper‑parameter and the selection criterion are reported (\yn{yes})
        \item Code for data preprocessing is included (\yn{yes} — scripts in supplementary ZIP; Dockerfile to be added post‑review)
        \item Source code needed to run experiments is included (\yn{yes} — private GitHub link will be made public upon acceptance)
        \item Code will be released publicly with a permissive license (\yn{yes} — MIT intended)
        \item Code includes comments with implementation details and references (\yn{yes} — comments will be finalized before release)
        \item Seed‑setting methods for stochastic algorithms are described (\yn{yes} — API generation uses deterministic decoding; seed = 42 documented in Experiments)
        \item Computing infrastructure (hardware/software specs) is reported (\yn{yes})
        \item Evaluation metrics are formally described with motivations (\yn{yes})
        \item Number of runs used for each reported result is specified (\yn{yes})
        \item Analysis of variation, confidence, or distributions is provided (NA — method deterministic)
        \item Significance of performance differences is assessed with statistical tests (NA)
        \item Final hyper‑parameter settings are listed (\yn{yes})
    \end{itemize}
\end{enumerate}

\clearpage            
\appendix

\FloatBarrier 
\begin{table*}[!t]
  \centering
  \setlength{\tabcolsep}{2pt}
  \renewcommand{\arraystretch}{0.8}
  \begin{tabularx}{\linewidth}{@{}lcccccX@{}}
    \toprule
    \textbf{Model} & \textbf{Params} & \textbf{Ctx Len} & \textbf{Max Out} & \textbf{Quant.} & \textbf{Licence} & \textbf{Spec Refs.} \\ \midrule
    \model{o4-mini}      & — & 128 K & 16 384 & n/a\textsuperscript{†} & Proprietary & Sys. card \cite{openai2025o4mini} \\ 
    \model{gpt-4o}       & — & 128 K & 4 096  & n/a\textsuperscript{†} & Proprietary & Sys. card \cite{openai2024gpt4o} \\ 
    \model{gpt-4o-mini}  & — & 128 K & 16 384 & n/a\textsuperscript{†} & Proprietary & API docs \cite{openai2025gpt4omini} \\ 
    \model{gpt-4.1}      & — &   1 M & 32 768 & n/a\textsuperscript{†} & Proprietary & API rel. \cite{openai2025gpt41} \\ 
    \model{gpt-4.1-mini} & — &   1 M & 32 768 & n/a\textsuperscript{†} & Proprietary & API rel. \cite{openai2025gpt41mini} \\ 
    \model{gpt-4.1-nano} & — &   1 M & 32 768 & n/a\textsuperscript{†} & Proprietary & API rel. \cite{openai2025gpt41nano} \\ \midrule
    DeepSeek-V3          & 671 B (37 B act.) & 128 K & 8 000  & n/a\textsuperscript{†} & DeepSeek Licence & Tech rep. \cite{deepseek2024v3} \\ 
    DeepSeek-R1          & 671 B (37 B act.) & 64 K  & 64 000 & n/a\textsuperscript{†} & DeepSeek Licence & Tech rep. \cite{deepseek2025r1} \\ \midrule
    Qwen3-4B             & 4 B  & 32 768 & 38 912 & FP16 & Apache 2.0 & Report \cite{qin2024qwen34b} \\ 
    Qwen3-8B             & 8 B  & 32 768 & 38 912 & FP16 & Apache 2.0 & Report \cite{qin2024qwen38b} \\ 
    Qwen3-32B-AWQ        & 32 B & 32 768 & 38 912 & 4-bit AWQ & Apache 2.0 & Report \cite{qin2024qwen332bawq} \\ 
    \bottomrule
  \end{tabularx}
  \caption{Comprehensive specifications for all evaluated models.  
           Quant.\ column shows the weight/compute precision **we actually used**.  
           \textsuperscript{†}\,For vendor-hosted APIs the underlying precision is managed server-side and not exposed to users; OpenAI and DeepSeek documentation state that mixed/half-precision inference (FP16 or BF16) is used internally, but no direct control is provided.}
  \label{tab:app_models}
\end{table*}

\section{Extended Model Specifications}
\label{app:models}

\FloatBarrier 
\begin{table*}[t]
\centering
\begin{tabular}{lrrrrrr}
\toprule
\multirow{2}{*}{Model}
& \multicolumn{2}{c}{MAE $\downarrow$}
& \multicolumn{2}{c}{MALD $\downarrow$}
& \multicolumn{2}{c}{EM $\uparrow$}\\
 & Baseline & CAPEL & Baseline & CAPEL & Baseline & CAPEL\\\midrule
  \texttt{o4-mini}         &   \textbf{34.42} & 347.08 & \textbf{0.04} & 0.48 & 41.6\% & \textbf{51.2\%}\\
  GPT-4.1                  &  129.17 &  \textbf{2.00} & 0.24 & \textbf{0.00} & 1.9\% & \textbf{94.2\%}\\
  GPT-4.1-mini &  486.31 &  \textbf{26.39} & 0.84 & \textbf{0.05} & 1.6\% & \textbf{38.9\%}\\
  GPT-4.1-nano             &   40.62 &   40.62 & 0.06 & 0.06 & 46.0\% & 46.0\%\\
  GPT-4o                   &  167.75 &  \textbf{87.30} & 0.59 & \textbf{0.11} & 1.3\% & \textbf{78.0\%}\\
  GPT-4o-mini              &  140.59 &  \textbf{6.68} & 0.28 & \textbf{0.01} & 2.3\% & \textbf{75.9\%}\\
  Qwen3-32B-AWQ            &  292.02 &  \textbf{57.77} & 0.49 & \textbf{0.07} & 0.6\% & \textbf{83.1\%}\\
  Qwen3-32B-AWQ (+think)   &  353.74 &  \textbf{332.63} & 0.68 & \textbf{0.64} & 5.6\% & \textbf{80.9\%}\\
  Qwen3-8B                 & 1101.35 &  \textbf{50.19} & 1.62 & \textbf{0.07} & 0.6\% & \textbf{72.1\%}\\
  Qwen3-4B                 &  728.60 &  \textbf{180.96} & 1.08 & \textbf{0.31} & 0.6\% & \textbf{49.7\%}\\
  Qwen3-4B (+think)        & 6653.68 &  \textbf{4341.71} & 16.06 & \textbf{13.51} & 1.8\% & \textbf{31.5\%}\\
  DeepSeek-V3            &  617.64 &  \textbf{8.33} & 1.47 & \textbf{0.01} & 0.7\% & \textbf{86.8\%}\\
  DeepSeek-R1            &  \textbf{825.00} &  - & \textbf{1.39} & - & \textbf{14.1\%} & -\\
\bottomrule
\end{tabular}
\caption{Random Text Generation - English}. 
\label{tab:rtg-en}
\end{table*}

\begin{table*}[t]
  \centering
\begin{tabular}{lrrrrrr}
\toprule
\multirow{2}{*}{Model}
& \multicolumn{2}{c}{MAE $\downarrow$}
& \multicolumn{2}{c}{MALD $\downarrow$}
& \multicolumn{2}{c}{EM $\uparrow$}\\
 & Baseline & CAPEL & Baseline & CAPEL & Baseline & CAPEL\\\midrule
  \texttt{o4-mini}         &   \textbf{36.41} & 652.88 & \textbf{0.04} & 0.76 & \textbf{72.3\%} & 18.0\%\\
  GPT-4.1                  &   96.32 &  \textbf{10.68} & 0.19 & \textbf{0.02} & 1.8\% & \textbf{53.7\%}\\
  GPT-4.1-mini ($T{=}1.0$) &  331.07 &  \textbf{13.34} & 0.57 & \textbf{0.02} & 0.6\% & \textbf{6.0\%}\\
  GPT-4.1-nano             &  568.84 &  \textbf{119.79} & 0.99 & \textbf{0.19} & 0.5\% & \textbf{9.0\%}\\
  GPT-4o                   &  175.58 &  \textbf{59.25} & 0.45 & \textbf{0.07} & 0.7\% & \textbf{21.0\%}\\
  GPT-4o-mini              &  \textbf{183.92} & 285.28 & 0.37 & \textbf{0.32} & 0.3\% & \textbf{1.0\%}\\
  Qwen3-32B-AWQ            &  386.95 &  \textbf{66.54} & 0.66 & \textbf{0.07} & 0.5\% & \textbf{71.0\%}\\
  Qwen3-32B-AWQ (+think)   &  \textbf{643.40} & 1896.46 & \textbf{1.36} & 6.39 & \textbf{10.1\%} & 0.0\%\\
  Qwen3-8B                 & 1348.32 &  \textbf{21.79} & 2.48 & \textbf{0.03} & 0.5\% & \textbf{78.0\%}\\
  Qwen3-8B (+think)        & 20941.63 &  \textbf{3663.44} & 61.32 & \textbf{10.22} & 3.9\% & \textbf{0.0\%}\\
  Qwen3-4B                 &  \textbf{1306.04} & 1599.02 & 3.07 & \textbf{2.03} & 0.3\% & \textbf{32.0\%}\\
  DeepSeek-V3            &  202.90 &  \textbf{122.74} & 0.35 & \textbf{0.18} & 0.5\% & \textbf{1.0\%}\\
  DeepSeek-R1            &  \textbf{79.43} &  - & \textbf{0.12} & - & \textbf{39.0\%} & -\\
\bottomrule
\end{tabular}
  \caption{\textbf{Random Text Generation – Chinese}.}
  \label{tab:rtg-zh}
\end{table*}

\section{Additional Random‑Text‑Generation Results (Chinese)}
\label{app:rtg-zh}


\FloatBarrier 
\begin{table*}[t]
\centering
\setlength{\tabcolsep}{3pt}
\begin{tabular}{lrrrrrrrrrrrr}
\toprule
\multirow{2}{*}{Model}
& \multicolumn{2}{c}{MAE $\downarrow$}
& \multicolumn{2}{c}{MALD $\downarrow$}
& \multicolumn{2}{c}{Exact Match $\uparrow$}
& \multicolumn{2}{c}{ROUGE-1}
& \multicolumn{2}{c}{ROUGE-2}
& \multicolumn{2}{c}{ROUGE-L}\\
\cmidrule(lr){2-3}\cmidrule(lr){4-5}\cmidrule(lr){6-7}\cmidrule(lr){8-9}\cmidrule(lr){10-11}\cmidrule(lr){12-13}
& Base & CAPEL & Base & CAPEL & Base & CAPEL & Base & CAPEL & Base & CAPEL & Base & CAPEL\\
\midrule
GPT-4.1                  &  1.02 &   0.04 & 0.049 & 0.002 & 0.304 & 0.960 & 0.2548 & 0.2458 & 0.0618 & 0.0601 & 0.1878 & 0.1788 \\
GPT-4.1-mini             &  1.58 &   0.28 & 0.077 & 0.012 & 0.172 & 0.740 & 0.2230 & 0.2167 & 0.0452 & 0.0404 & 0.1643 & 0.1527 \\
GPT-4.1-nano             &  1.32 &   0.27 & 0.063 & 0.012 & 0.249 & 0.855 & 0.2224 & 0.1851 & 0.0427 & 0.0222 & 0.1624 & 0.1325 \\
DeepSeek-V3              &  2.99 &   0.05 & 0.138 & 0.002 & 0.091 & 0.963 & 0.2413 & 0.2182 & 0.0545 & 0.0389 & 0.1760 & 0.1563 \\
Qwen3-4B                 &  4.12 &   0.21 & 0.194 & 0.008 & 0.068 & 0.958 & 0.2350 & 0.1654 & 0.0468 & 0.0224 & 0.1740 & 0.1215 \\
Qwen3-8B                 &  6.19 &   0.15 & 0.282 & 0.008 & 0.071 & 0.956 & 0.2481 & 0.1757 & 0.0520 & 0.0252 & 0.1766 & 0.1292 \\
\bottomrule
\end{tabular}
\caption{XSUM performance under reference-length budget.}
\label{tab:xsum-ref}
\end{table*}

\begin{table*}[t]
\centering
\setlength{\tabcolsep}{3pt}
\begin{tabular}{lrrrrrrrrrrrr}
\toprule
\multirow{2}{*}{Model}
& \multicolumn{2}{c}{MAE $\downarrow$}
& \multicolumn{2}{c}{MALD $\downarrow$}
& \multicolumn{2}{c}{Exact Match $\uparrow$}
& \multicolumn{2}{c}{ROUGE-1}
& \multicolumn{2}{c}{ROUGE-2}
& \multicolumn{2}{c}{ROUGE-L}\\
\cmidrule(lr){2-3}\cmidrule(lr){4-5}\cmidrule(lr){6-7}\cmidrule(lr){8-9}\cmidrule(lr){10-11}\cmidrule(lr){12-13}
& Base & CAPEL & Base & CAPEL & Base & CAPEL & Base & CAPEL & Base & CAPEL & Base & CAPEL\\
\midrule
GPT-4.1                  &  1.78 &   0.46 & 0.036 & 0.009 & 0.181 & 0.858 & 0.2402 & 0.2126 & 0.0585 & 0.0464 & 0.1603 & 0.1429 \\
GPT-4.1-mini             &  3.10 &   1.15 & 0.062 & 0.023 & 0.076 & 0.417 & 0.2201 & 0.1996 & 0.0447 & 0.0366 & 0.1459 & 0.1303 \\
GPT-4.1-nano             &  3.15 &   1.96 & 0.063 & 0.039 & 0.087 & 0.650 & 0.2170 & 0.1926 & 0.0413 & 0.0281 & 0.1434 & 0.1252 \\
Qwen3-32B-AWQ            &  5.85 &  11.81 & 0.117 & 0.236 & 0.049 & 0.663 & 0.2413 & 0.2063 & 0.0535 & 0.0395 & 0.1612 & 0.1352 \\
Qwen3-4B                 &  6.59 & 109.12 & 0.132 & 2.182 & 0.048 & 0.909 & 0.2075 & 0.1344 & 0.0360 & 0.0186 & 0.1374 & 0.0944 \\
Qwen3-8B                 &  6.56 & 111.26 & 0.131 & 2.225 & 0.050 & 0.908 & 0.2039 & 0.1396 & 0.0368 & 0.0199 & 0.1338 & 0.0974 \\
\bottomrule
\end{tabular}
\caption{XSUM performance under 50-word budget.}
\label{tab:xsum-50}
\end{table*}

\begin{table*}[t]
\centering
\setlength{\tabcolsep}{3pt}
\begin{tabular}{lrrrrrrrrrrrr}
\toprule
\multirow{2}{*}{Model}
& \multicolumn{2}{c}{MAE $\downarrow$}
& \multicolumn{2}{c}{MALD $\downarrow$}
& \multicolumn{2}{c}{Exact Match $\uparrow$}
& \multicolumn{2}{c}{ROUGE-1}
& \multicolumn{2}{c}{ROUGE-2}
& \multicolumn{2}{c}{ROUGE-L}\\
\cmidrule(lr){2-3}\cmidrule(lr){4-5}\cmidrule(lr){6-7}\cmidrule(lr){8-9}\cmidrule(lr){10-11}\cmidrule(lr){12-13}
& Base & CAPEL & Base & CAPEL & Base & CAPEL & Base & CAPEL & Base & CAPEL & Base & CAPEL\\
\midrule
GPT-4.1                  &  3.54 &   0.33 & 0.030 & 0.003 & 0.098 & 0.815 & 0.1686 & 0.1601 & 0.0424 & 0.0365 & 0.1101 & 0.1032 \\
GPT-4.1-mini             &  5.50 &   5.81 & 0.046 & 0.048 & 0.057 & 0.319 & 0.1571 & 0.1478 & 0.0355 & 0.0291 & 0.1024 & 0.0948 \\
GPT-4.1-nano             &  6.10 &   4.78 & 0.051 & 0.040 & 0.057 & 0.520 & 0.1541 & 0.1537 & 0.0324 & 0.0287 & 0.1002 & 0.1000 \\
Qwen3-4B                 & 31.67 &  28.50 & 0.264 & 0.238 & 0.000 & 0.517 & 0.1811 & 0.1212 & 0.0374 & 0.0202 & 0.1191 & 0.0824 \\
Qwen3-8B                 & 12.61 &  11.24 & 0.105 & 0.094 & 0.028 & 0.760 & 0.1646 & 0.1358 & 0.0373 & 0.0252 & 0.1088 & 0.0902 \\
\bottomrule
\end{tabular}
\caption{XSUM performance under 120-word budget.}
\label{tab:xsum-120}
\end{table*}

\section{Additional \textsc{XSUM} Fixed Target Length Results}
\label{app:xsum}


\FloatBarrier 

\begin{table*}[t]
  \centering
  \setlength{\tabcolsep}{4pt}
  \begin{tabular*}{\textwidth}{@{\extracolsep{\fill}} llrrrr}
    \toprule
    \textbf{Model} & \textbf{Method}
      & \textbf{MALD}$\downarrow$ & \textbf{EM\,(\%)}$\uparrow$
      & \textbf{Avg Iter} & \textbf{Score}$\uparrow$ \\
    \midrule
    \multirow{3}{*}{gpt-4.1}%
      & Acc      & 0.135 & 32.0 & 11.62 & 4.23 \\
      & AccMem   & \textbf{0.013} & 62.9 & \textbf{8.53} & 5.78 \\
      & Mem      & 0.018 & \textbf{67.4} & 8.25 & 5.78 \\
    \midrule
    \multirow{3}{*}{gpt-4.1-mini}%
      & Acc      & 0.087 & 26.9 & 11.74 & 3.82 \\
      & AccMem   & 0.045 & 46.3 & 9.90 & 6.30 \\
      & Mem      & 0.041 & 44.6 & 9.83 & 3.74 \\
    \midrule
    \multirow{3}{*}{gpt-4.1-nano}%
      & Acc      & 0.161 & 15.4 & 13.38 & 4.96 \\
      & AccMem   & 0.058 & 21.7 & 12.97 & 5.27 \\
      & Mem      & 0.060 & 24.0 & 12.83 & 4.12 \\
    \midrule
    o4-mini & Acc & 0.031 & 87.4 & 2.21 & 5.14 \\
    \bottomrule
  \end{tabular*}
  \caption{BB-MH decoding results on MT-Bench-LI.  
    Methods are abbreviated as Acc=iterative\_acceptance,  
    AccMem=iterative\_acceptance\_memory, and Mem=iterative\_memory.  
    Avg Iter shows the average number of refinement iterations (max 15).  
    The “Score” column refers to the single-answer MT-Bench scores.}
  \label{tab:mtbench-bbmh}
\end{table*}

\section{\textsc{BB-MH} Results}
\label{app:bbmh}

\paragraph{Method.}
We reproduce the three Metropolis–Hastings (MH) variants
(\textit{Acc}, \textit{Mem}, and \textit{AccMem}) described in
§\ref{sec:mtbench-results}.  
Hyper‑parameters follow \citet{Gu2024BlackBox} exactly, except that we
enforce the strict equality filter $|y|{=}T$ required by MT‑Bench‑LI.

\subsection{Iterative versus One‑Shot Control}
\label{app:iter-vs-one-shot}

Table~\ref{tab:mtbench-bbmh} shows that the
\textit{iterative\_acceptance\_memory} variant generally yields the best
Exact Match among the MH‑based methods (e.g., \textbf{92.9 \%} on
\texttt{gpt‑4.1}) but still trails the one‑shot \textit{CAPEL} prompt by
3–8 percentage points while incurring \emph{eight to ten} additional
model calls on average.  
Surprisingly, removing the MH filter (\textit{iterative\_memory}) often
matches or \emph{exceeds} \textit{iterative\_acceptance} in both EM and
MAE, suggesting that history retention contributes more than the
acceptance criterion when $k{=}1$.  
Conversely, the lightweight \texttt{o4-mini} model attains 87.4 \% EM
with only $\approx2$ iterations in the
\textit{iterative\_acceptance} setting, underscoring that smaller yet
reasoning‑focused models can benefit more from local rewriting than
their larger counterparts.

\FloatBarrier 

\begin{figure*}[t]
  \centering
  \includegraphics[width=\textwidth]{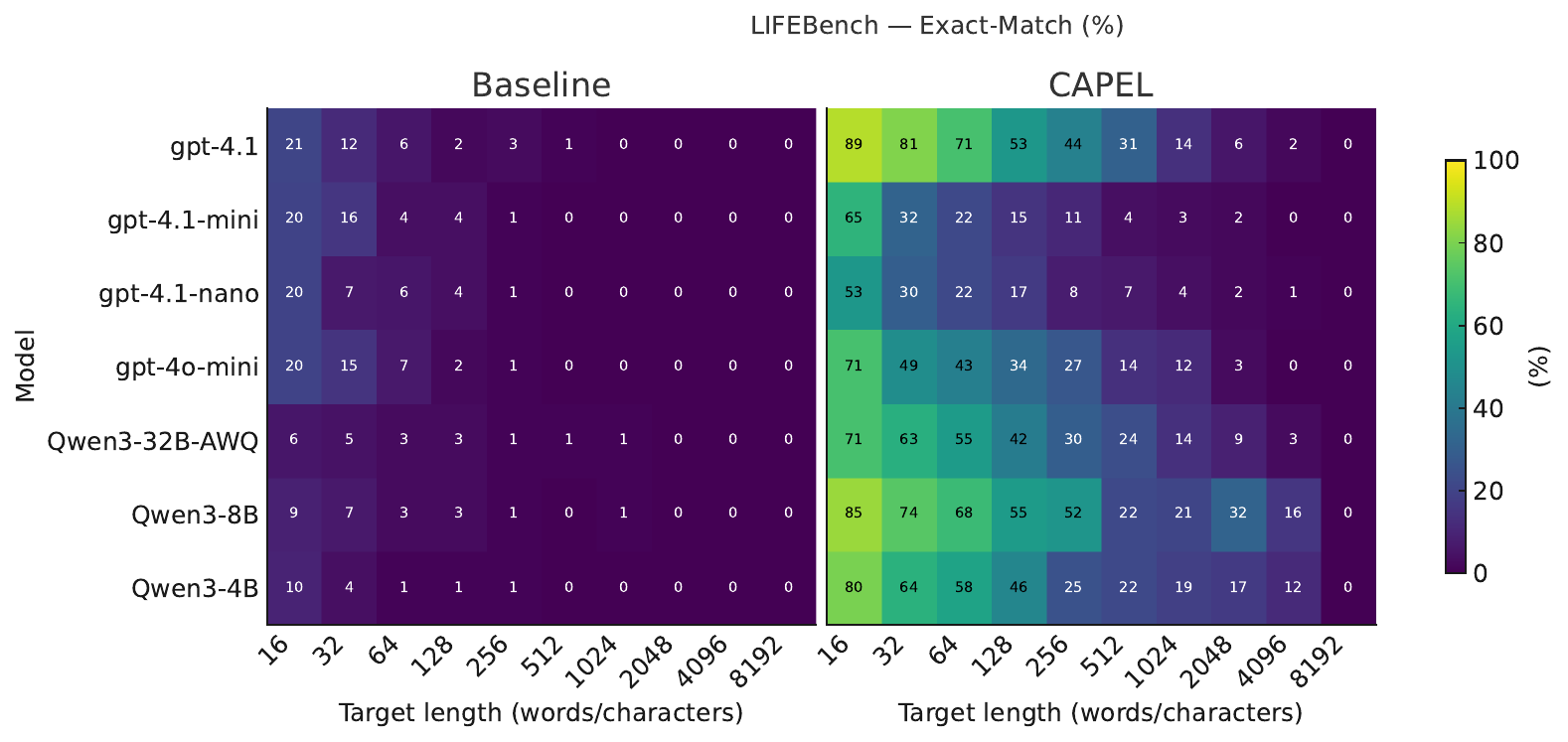}
  \caption{Exact-match rate (\%) on \textsc{LIFEBench}.
  Left = \textsc{Baseline}, right = \textsc{CAPEL}.  
  X-axis = target length (\emph{words/characters});
  Y-axis = model.}
  \label{fig:lifebench-em}
\end{figure*}

\section{Additional \textsc{LIFEBench} Exact‑Match Results}
\label{app:lifebench}


\FloatBarrier 

\begin{figure*}[t]
\centering
\includegraphics[width=\textwidth]{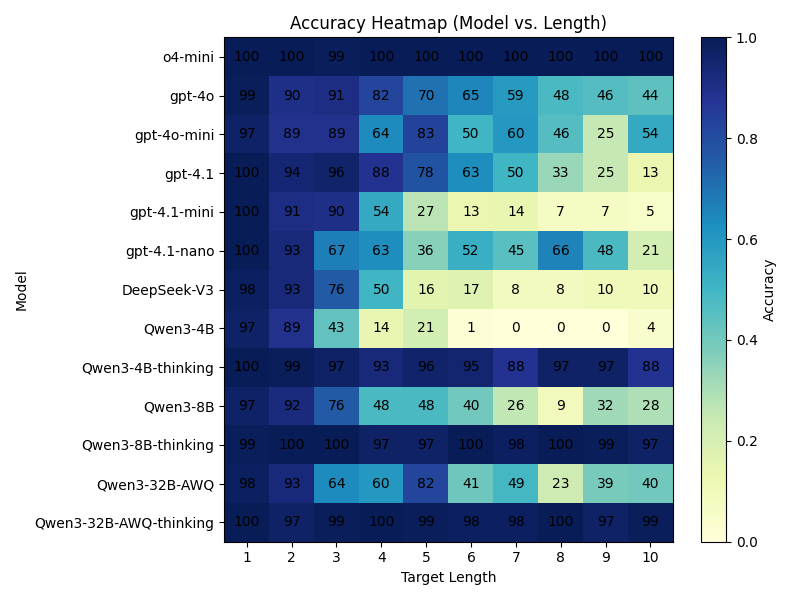}
\caption{Exact‑match accuracy heatmap for the token‑counting diagnostic on
English sentences (word level).  Rows correspond to eleven evaluated LLMs,
columns to target lengths $L=1$--$10$.  Warmer colors indicate higher accuracy.}
\label{fig:counting-heatmap-en}
\end{figure*}

\begin{figure*}[t]
\centering
\includegraphics[width=\textwidth]{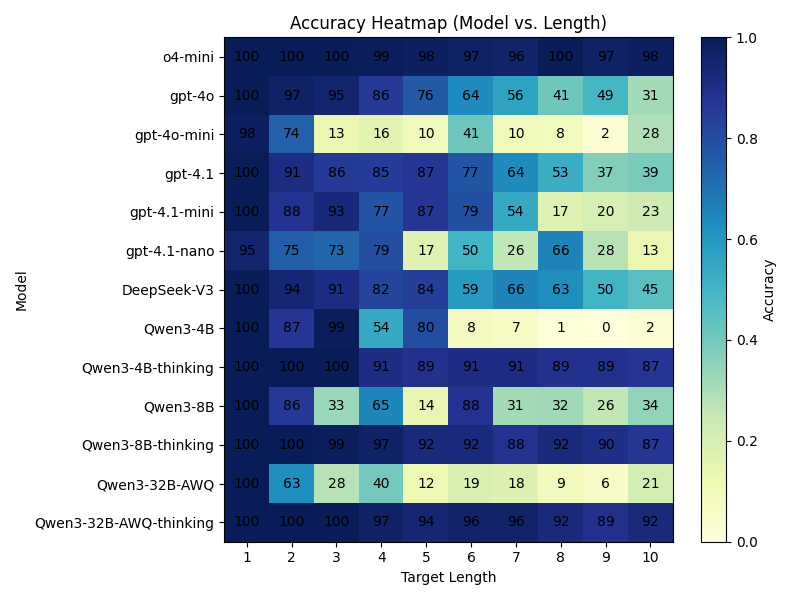}
\caption{Exact‑match accuracy heatmap for the token‑counting diagnostic on
Chinese sentences (character level).  Rows correspond to eleven evaluated LLMs,
columns to target lengths $L=1$--$10$.  Warmer colors indicate higher accuracy.}
\label{fig:counting-heatmap-zh}
\end{figure*}

\section{Additional Figures for the Counting Diagnostic}
\label{sec:appendix-count}

\end{document}